%% file: main.tex
\documentclass[journal]{IEEEtran}
\usepackage{graphicx}
\usepackage{tikz}
\usepackage{amsfonts}
\usepackage{caption}
\usepackage{subcaption}
\usetikzlibrary{decorations.pathreplacing,angles,quotes,positioning,shapes,arrows}

\newcommand{\aainp}[0]{\mathbb{R}^{3 \times 3 \times 20}}

\ifCLASSINFOpdf
\else
\fi
\graphicspath{ {pictures/} }

\begin{document}
%
\title{Computational Graph Approach for Detection of Composite Human Activities}
%
%
%

\author{
    Niko~Reunanen,
    Ville~K{\"o}n{\"o}nen,
    Hermanni~H{\"a}lv{\"a},
    Jani~M{\"a}ntyj{\"a}rvi,
    Arttu L{\"a}ms{\"a} and 
    Jussi Liikka

    \thanks{N. Reunanen, V. K{\"o}n{\"o}nen, J. M{\"a}ntyj{\"a}rvi, A. L{\"a}ms{\"a} and J. Liikka are with the VTT Technical Research Centre of Finland, Kaitov{\"a}yl{\"a} 1, 90571, Oulu, Finland. E-mail: niko.reunanen@vtt.fi, ville.kononen@vtt.fi, jani.mantyjarvi@vtt.fi, arttu.lamsa@vtt.fi, jussi.liikka@vtt.fi.}%
    \thanks{H. H{\"a}lv{\"a} is with the VTT Technical Research Centre of Finland, Tekniikantie 1, 02150, Espoo, Finland. E-mail: hermanni.halva@vtt.fi.}%
}

%
%

\markboth{VTT Technical Research Centre of Finland Research Note}%
{}
%



\maketitle

\begin{abstract}

Existing work in human activity detection classifies physical activities using a single fixed-length subset of a sensor signal. However, temporally consecutive subsets of a sensor signal are not utilized. This is not optimal for classifying physical activities (composite activities) that are composed of a temporal series of simpler activities (atomic activities). A sport consists of physical activities combined in a fashion unique to that sport. The constituent physical activities and the sport are not fundamentally different. We propose a computational graph architecture for human activity detection based on the readings of a triaxial accelerometer. The resulting model learns 1) a representation of the atomic activities of a sport and 2) to classify physical activities as compositions of the atomic activities. The proposed model, alongside with a set of baseline models, was tested for a simultaneous classification of eight physical activities (walking, Nordic walking, running, soccer, rowing, bicycling, exercise bicycling and lying down). The proposed model obtained an overall mean accuracy of $77.91$\% (population) and $95.28$\% (personalized). The corresponding accuracies of the best baseline model were $73.52$\% and $90.03$\%. However, without combining consecutive atomic activities, the corresponding accuracies of the proposed model were $71.52$\% and $91.22$\%. The results show that our proposed model is accurate, outperforms the baseline models and learns to combine simple activities into complex activities. Composite activities can be classified as combinations of atomic activities. Our proposed architecture is a basis for accurate models in human activity detection.

\end{abstract}



%
\IEEEpeerreviewmaketitle

\input{./sections/1st.tex}

\input{./sections/2nd.tex}

\input{./sections/3rd.tex}

\input{./sections/4th.tex}

\ifCLASSOPTIONcaptionsoff
  \newpage
\fi



%

\bibliographystyle{IEEEtran}
\bibliography{references}

\end{document}

%% file: sections/1st.tex
\section{Introduction}
Human activity detection is the process of classifying the current physical activity of a person. A common data source for human activity detection is instantaneous acceleration (used in \cite{973004,mathie-classify-daily-movements,Karantonis:2006:IRH:2222932.2223358,6780615,Khan:2010:TAP:1870372.1870376,chernbumroong-activity,godfrey-activity-class,Kwapisz:2011:ARU:1964897.1964918}). Accelerometers are lightweight \cite{6217313}, inexpensive \cite{6399499}, and widely available \cite{6346377} across wrist devices and smart phones \cite{Kwapisz:2011:ARU:1964897.1964918}, and thus any built-in system could be distributed broadly.



Arguably, some physical activities are more complex than others in that they consist of other, more fundamental, sub-activities. We introduce the following terms: composite activities and atomic activities. Each composite activity is a unique temporal series of consecutive atomic activities. Therefore the atomic activities are building blocks for defining complex activities. Some sports are complex activities (e.g. basketball) that are combinations of constituent physical activities (atomic activities). Notice that simple activities are also defined as composite activities in which consecutive atomic activities are almost identical.

Previous work in human activity detection typically consists of the following steps: aggregation of a subset of a sensor signal, summarization of the information in the subset (e.g. sample statistics) and an instantaneous classification of the physical activity (used in \cite{6780615,Kwapisz:2011:ARU:1964897.1964918,Parkka:2006,Ermes:2008:DDA:2222942.2223513}). Models that utilize the previous three steps face a fundamental problem in classifying composite activities. For instance, the act of playing basketball includes running, and thus these models may struggle to learn separate representations for them. Our proposed model enables the detection of such complex physical activities by first decomposing them into unique atomic activities (Section \ref{sec:fe_at}) and then combining these for activity classification (Section \ref{sec:fe_co}).


In this paper, we propose a model for classifying the physical activity of a person based on the readings of a triaxial accelerometer. Compared to the existing work, our model has the following benefits:

\begin{itemize}
    \item The model learns to classify the physical activities as composite activities, which consist of atomic activities.
    \item The computed features, which characterize the atomic activities and the composite activities, are not selected from a fixed set of features (e.g. mean or variance, as in \cite{6780615,Kwapisz:2011:ARU:1964897.1964918}). Instead, the features are learned automatically to represent the supported physical activities. Therefore we do not impose assumptions on how to build a representation of the subsets of a sensor signal to summarize their information.
    \item The model can be utilized as a population system and as a personalized system. For a test subject, a population system is created using data of other people and a personalized system is created using data of the test subject. Our proposed model is used to experiment with both systems in Section \ref{sec:experiments}.
\end{itemize}

The proposed model is trained end-to-end as a single model without manual intervention. The computational components of our model originate from deep learning literature \cite{Schmidhuber201585}. However, the model uses a limited amount of memory and is not a deep model. Instead, one should consider our model to be a computational graph \cite{Koller:2009:PGM:1795555,Bastien-Theano-2012}.



The experiments in this paper use a total of six hours of data from Palantir Context Data Library \cite{Parkka:2006,Ermes:2008:DDA:2222942.2223513}. This data was created by recording multiple sensors, including a triaxial wrist-worn accelerometer, while users performed various activities. The sampling frequency of the accelerometer was 20Hz, which is sufficient to detect physical activities \cite{6780615}. The data were recorded outside a laboratory in a real-world environment, and represent a set of activities, which are performed in personal ways (e.g. two persons run in different ways.) Therefore, by utilizing the Palantir data, we obtain a good estimate of the real-world accuracy of our model for human activity detection. See \cite{Parkka:2006,Ermes:2008:DDA:2222942.2223513} for a detailed description of the data and the setup of the data collection.


The scope of our work in this paper is in the classification of eight physical activities (walking, Nordic walking, running, soccer, rowing, bicycling, exercise bicycling and lying down). A physical activity is classified using a subset of 15 seconds of acceleration signal. We also construct and evaluate a set of baseline models to classify the 15s of acceleration signal. The following established classifiers are selected as the baseline models in the experiments: logistic regression \cite{bishop}, random forest classifier \cite{Breiman:2001:RF:570181.570182}, decision tree \cite{bishop}, adaptive boosting classifier \cite{Freund+Schapire:1996} and linear support vector machine \cite{bishop}. The proposed model obtained an overall mean accuracy of $77.91$\% (population) and $95.28$\% (personalized). The corresponding accuracies of the best baseline model (random forest) were $73.52$\% and $90.03$\%.

The reported parametrization of our proposed model in this paper (e.g. the number of computational units) was selected through empirical experiments. Our goal is not to provide a universal and general parametrization for the model. However, the parametrization utilized in this paper is a solid initial choice, which is experimentally validated. Rather, our goal is to introduce a suitable architecture for a model to detect composite activities. The following section defines the architecture and its parametrization used in this paper.

%% file: sections/2nd.tex
\section{The proposed model}
To detect composite activities, we propose a model with three major components:
\begin{enumerate}
    \item Feature learning for atomic activities (Section \ref{sec:fe_at}).
    \item Feature learning for a composite activity (Section \ref{sec:fe_co}).
    \item Classification of the composite activity (Section \ref{sec:fe_cla}).
\end{enumerate}


An atomic activity is encoded into a vector of real values, which encode its characteristics. These values are called features and the process of learning to compute them is called feature learning (see Section \ref{sec:fe_at}). A set of feature vectors, corresponding to atomic activities, is sent to a recurrent model which is used as it is able to learn the temporal relationship between these features \cite{Hochreiter:1997:LSM:1246443.1246450}. The final internal state of the recurrent model, which is a vector of real values, represents the feature values of the composite activity. Therefore the representation of a composite activity is learned using the atomic activities (see Section \ref{sec:fe_co}). The final internal state is then classified into one of the supported physical activities (see Section \ref{sec:fe_cla}). The following subsection introduces the computational machinery, which are then examined in further detail in sections \ref{sec:fe_at}, \ref{sec:fe_co} and \ref{sec:fe_cla}.

\subsection{Computation in layers}
We factor the computation of our proposed model into consecutive steps called layers. A layer takes an input, transforms the input and outputs the transformed value. A set of chained layers forms a computational graph. For example, neural networks can be defined as computational graphs. The following layers are utilized by our model (see \cite{Schmidhuber201585,Hochreiter:1997:LSM:1246443.1246450,deep_learning_nature_hinton,NIPS2012_4824,Bengio2012} for more details):


\begin{itemize}
    \item Fully connected layer (FC). The layer applies a non-linear function element-wise to an affine transformation: $\sigma(\mathbf{W}\mathbf{x} + \mathbf{b})$ where $\mathbf{x} \in \mathbb{R}^d$ denotes $d$-dimensional input and $\mathbf{W}, \mathbf{b}$ are adjustable parameters. We utilize a rectified linear unit for the non-linearity ($\sigma(a) = max(0, a)$), which is widely employed as the non-linearity \cite{deep_learning_nature_hinton} because of its calculational simplicity \cite{AISTATS2011_GlorotBB11}. We use fully connected layers to learn suitable features to classify the physical activities.
    \item Convolution layer (CO). The layer processes an input signal while retaining its temporal structure. Similarly to a FC layer, a convolution layer applies affine transformations followed by a non-linearity. The parameters for affine transformations are contained in convolution kernels. The same convolution kernel parameters are used across all the temporal subsets of the input signal. Each convolution kernel in each layer outputs a new transformed signal. Convolution layers can use multiple convolution kernels to output multiple transformed signals. The goal is to learn translation invariant features (kernels) from the input signal.
    \item Max-pooling layer (MP). This layer reduces the dimensionality of its input signal by outputting only the largest input value from temporally neighboring regions. The size of a region is four without overlapping in our experiments. Therefore, for example, eight temporally consecutive acceleration readings are transformed into two values.
    \item Long short term memory layer (LSTM). LSTM is a specific type of a recurrent neural network, which utilizes a gating mechanism to retain an internal state over a long period of time in its memory \cite{Hochreiter:1997:LSM:1246443.1246450}. During every time step, the LSTM determines its internal state based on the current input vector (atomic feature vector) and its own previous state. Each composite activity is modeled using the final state of the LSTM.
    \item Softmax layer (SM). The layer computes a probability distribution (relative probabilities) for the physical activities as $P(i) = e^{z_i}(\sum\nolimits^8_{j=1}e^{z_j})^{-1}$ where $P(i)$ is the probability of $i$:th activity and $z_i$ is the $i$:th input value for the layer. We use a softmax layer, and the resulting probability distribution, to predict the current physical activity.
\end{itemize}


The layers compute the gradient of their parameters with respect to its output (e.g. $\frac{\partial \sigma}{\partial \mathbf{W}} = \frac{\partial \sigma}{\partial a}\frac{\partial a}{\partial \mathbf{W}}$, $\frac{\partial \sigma}{\partial \mathbf{b}} = \frac{\partial \sigma}{\partial a}\frac{\partial a}{\partial \mathbf{b}}$ and $\frac{\partial \sigma}{\partial \mathbf{x}} = \frac{\partial \sigma}{\partial a}\frac{\partial a}{\partial \mathbf{x}}$ for FC). The gradient between consecutive layers is computed using the chain rule where the gradient of the lower layer is multiplied by the gradient of the upper layer. This allows one to compute the gradient of all parameters with respect to the desired output of the computational graph.

The deviation between the desired output ($y$) and the output of the computational graph ($\hat{y}$) is measured using an error function ($e(y,\hat{y})$). The final layer in our model is a softmax layer, which forms a probability distribution of the classification results. Therefore we obtain the probabilities of the eight physical activities. The classification result is the physical activity with the highest probability. The error function is the cross-entropy function between the computed probability distribution and the desired probability distribution, see \cite{Theodoridis:2008:PRF:1457541} for details. To minimize the error function, the parameter values are updated using gradient descent as:

\begin{equation}
    \mathbf{\Theta} = \mathbf{\Theta} - \alpha * \frac{\partial e}{\partial \mathbf{\Theta}} ~~~ ,
    \label{eq:grad_desc}
\end{equation}

where $\mathbf{\Theta}$ is the set of model parameters and $\alpha \in \mathbb{R}$ is a hyperparameter called learning rate. The learning rate scales the magnitude of the gradient descent \cite{bishop}. It is important to select an appropriate value of the learning rate for the convergence of the model. Section \ref{sec:prop_train} explains the training procedure of our proposed model in detail. The computational graph is trained by minimizing the error function. This approach is known as the backpropagation algorithm in the neural network literature, see \cite{Theodoridis:2008:PRF:1457541,bishop,Mitchell:1997:ML:541177} for more details. The following subsection describes the feature learning for atomic activities in detail.

\tikzstyle{block2_ell} = [ellipse, draw, fill=white!20, text width=3.0em, text centered, minimum height=1.8em]
\tikzstyle{block2} = [rectangle, draw, fill=white!20, text width=3.0em, text centered, minimum height=1.8em]
\tikzstyle{line} = [draw, -latex']

\subsection{Feature learning for atomic activities\label{sec:fe_at}}
An atomic activity is represented as a $d$-dimensional vector of real values ($\mathbf{x} \in \mathbb{R}^d$). We selected $d=128$ based on empirical tests using the values $d \in \{8,16,32,64,...,512\}$. The input for computing $\mathbf{x}$ in our experiments is three seconds of sensor signal from a triaxial accelerometer (three axes, 20Hz sampling rate.) Feature learning ($f: \aainp \longrightarrow \mathbb{R}^d$) transforms the acceleration signal into a feature representation of a fixed size ($d$).

\begin{figure}[!h]
    \begin{tikzpicture}[node distance = 1.0cm, auto]
    
        \node [block2_ell, text width=1.9em] (fe1) {Input};
        \node [block2, below of=fe1] (fe2) {CO1};
        \node [block2, below of=fe2] (fe3) {CO2};
        \node [block2, below of=fe3] (fe4) {MP};
        \node [block2, below of=fe4] (fe5) {FC};
        
        \node [below of=fe5] (finalss) {$\mathbf{x} \in \mathbb{R}^d$};
        \path [line] (fe5) -- (finalss);

        \node [block2_ell, left=0.8cm of fe1, text width=1.9em] (fe1_dup) {Input};
        \node [block2, below left=-0.1cm and 0.8cm of fe3] (pah) {AFL};

        \path [line] (fe1) -- (fe2);
        \path [line] (fe2) -- (fe3);
        \path [line] (fe3) -- (fe4);
        \path [line] (fe4) -- (fe5);
        \path [line] (fe1_dup) -- (pah);
        
        \draw[decoration={brace,mirror,raise=6pt,amplitude=10pt},decorate,line width=0.7pt] (fe2.north west) -- (fe5.south west) node [] {};
        
        \node [right=0.5cm of fe1, text width=7cm] {Acceleration signal (3s, $\aainp$)};
        \node [right=0.5cm of fe2, text width=7cm] {8 convolution kernels};
        \node [right=0.5cm of fe3, text width=7cm] {32 convolution kernels};
        \node [right=0.5cm of fe4, text width=7cm] {Max pooling of region size 4};
        \node [right=0.5cm of fe5, text width=7cm] {Fully connected, 128 outputs};

    \end{tikzpicture}
    \caption{\footnotesize An illustration of a computational graph for the feature learning for atomic activities (AFL). The features are encoded as a vector of 128 values ($\mathbf{x} \in \mathbb{R}^d, d=128$), which are computed by the fully connected layer (FC). The input data ($\aainp$) contain three seconds of acceleration signal from a triaxial accelerometer with a sampling rate of 20Hz.}
    \label{fig:feature_extraction_block}
\end{figure}
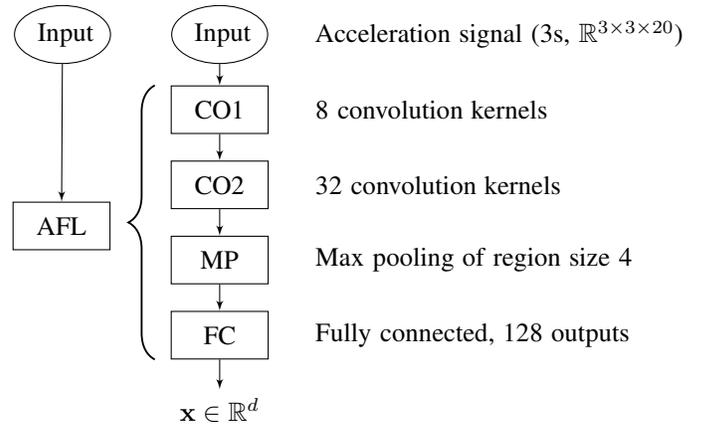

Fig. \ref{fig:feature_extraction_block} illustrates the architecture of the computational graph for learning and computing the features of an atomic activity (AFL). The computation of the features commences by processing the acceleration signal in two consecutive convolution layers. The convolution kernels learn to identify signal patterns that correspond to different atomic activities. We experimented using a various number of consecutive convolution layers (architecture) and kernels (parametrization). We utilize two layers with $8$ and $32$ kernels because they provided good results consistently. The maximum values of the twice convolved signal are pooled to reduce the number of parameters. The final layer is a fully connected layer, which acquires the pooled signal. The outputs of the fully connected layer are the feature values ($\mathbf{x}$) of an atomic activity. A larger model (see Fig. \ref{fig:comp_feat_ext}) utilizes these feature values (Fig. \ref{fig:feature_extraction_block}) to learn a composite activity from atomic activities. The following subsection describes the feature learning for composite activities in detail.

\subsection{Feature learning for composite activities\label{sec:fe_co}}
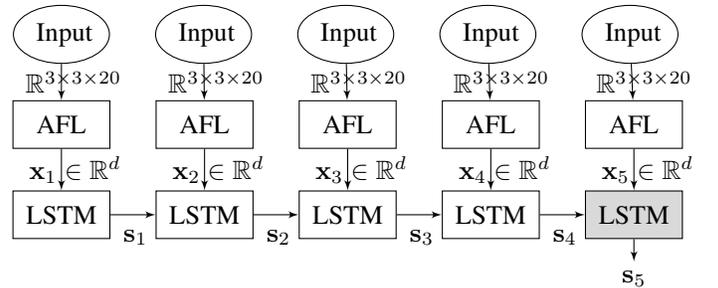
\begin{figure}[!h]
    \centering
    \begin{tikzpicture}[node distance = 1.2cm, auto]
    
        \node [block2_ell, text width=1.9em] (in1) {Input};
        \node [block2_ell, right=0.6cm of in1, text width=1.9em] (in2) {Input};
        \node [block2_ell, right=0.6cm of in2, text width=1.9em] (in3) {Input};
        \node [block2_ell, right=0.6cm of in3, text width=1.9em] (in4) {Input};
        \node [block2_ell, right=0.6cm of in4, text width=1.9em] (in5) {Input};

        \node [block2, below of=in1] (fe1) {AFL};
        \node [block2, right=0.6cm of fe1] (fe2) {AFL};
        \node [block2, right=0.6cm of fe2] (fe3) {AFL};
        \node [block2, right=0.6cm of fe3] (fe4) {AFL};
        \node [block2, right=0.6cm of fe4] (fe5) {AFL};

        \node [block2, below of=fe1] (lstm1) {LSTM};
        \node [block2, right=0.6cm of lstm1] (lstm2) {LSTM};
        \node [block2, right=0.6cm of lstm2] (lstm3) {LSTM};
        \node [block2, right=0.6cm of lstm3] (lstm4) {LSTM};
        \node [block2, right=0.6cm of lstm4, fill=gray!30] (lstm5) {LSTM};

        \path [line] (fe1) -- node[xshift=-0.55cm]{$\mathbf{x}_1 \in \mathbb{R}^d$} (lstm1);
        \path [line] (fe2) -- node[xshift=-0.55cm]{$\mathbf{x}_2 \in \mathbb{R}^d$} (lstm2);
        \path [line] (fe3) -- node[xshift=-0.55cm]{$\mathbf{x}_3 \in \mathbb{R}^d$} (lstm3);
        \path [line] (fe4) -- node[xshift=-0.55cm]{$\mathbf{x}_4 \in \mathbb{R}^d$} (lstm4);
        \path [line] (fe5) -- node[xshift=-0.55cm]{$\mathbf{x}_5 \in \mathbb{R}^d$} (lstm5);

        \path [line] (lstm1) -- (lstm2);
        \path [line] (lstm2) -- (lstm3);
        \path [line] (lstm3) -- (lstm4);
        \path [line] (lstm4) -- (lstm5);

        \path [line] (in1) -- node[xshift=-0.6cm]{$\aainp$} (fe1);
        \path [line] (in2) -- node[xshift=-0.6cm]{$\aainp$} (fe2);
        \path [line] (in3) -- node[xshift=-0.6cm]{$\aainp$} (fe3);
        \path [line] (in4) -- node[xshift=-0.6cm]{$\aainp$} (fe4);
        \path [line] (in5) -- node[xshift=-0.6cm]{$\aainp$} (fe5);

        \node [right=0.0cm of lstm1,xshift=0.05cm,yshift=-0.3cm] {$\mathbf{s}_1$};
        \node [right=0.0cm of lstm2,xshift=0.05cm,yshift=-0.3cm] {$\mathbf{s}_2$};
        \node [right=0.0cm of lstm3,xshift=0.05cm,yshift=-0.3cm] {$\mathbf{s}_3$};
        \node [right=0.0cm of lstm4,xshift=0.05cm,yshift=-0.3cm] {$\mathbf{s}_4$};
        
        \node [below=0.3cm of lstm5] (finalss) {$\mathbf{s}_5$};
        \path [line] (lstm5) -- (finalss);
        
    \end{tikzpicture}
    \caption{\footnotesize A composite activity is a temporal series of consecutive atomic activities, which are represented by feature values ($\mathbf{x}^d_i$). Long Short Term Memory obtains $\mathbf{x}^d_i$ chronologically one at a time to learn the temporal relationships between the atomic activities. The internal states of the LSTM ($\mathbf{s}_i$) are a representation of a composite activity, which is updated internally (arrows between LSTM blocks) and by the atomic activities (arrows from feature learning). The final internal state of the LSTM ($\mathbf{s}_5$) is the feature representation of the composite activity.}
    \label{fig:comp_feat_ext}
\end{figure}

In our experiments, a composite activity is segmented into five consecutive atomic activities, which results in a total of 15 seconds of acceleration signal. Using the feature learning for atomic activities (AFL) in Fig. \ref{fig:feature_extraction_block}, the atomic activities are transformed into five vectors of feature values ($\mathbf{x}_1, \mathbf{x}_2, \mathbf{x}_3, \mathbf{x}_4, \mathbf{x}_5$). The parametrization of AFL is shared across the temporally consecutive subsets (3s) of the acceleration signal. The feature vectors are provided to LSTM, one vector at a time in temporal order. The final internal state of LSTM ($\mathbf{s}_5$) is the feature representation of the composite activity, which encodes the composite activity using the atomic activities. Fig. \ref{fig:comp_feat_ext} illustrates the feature learning for composite activities. The following subsection describes how to classify the current physical activity of a person given the final internal state of the LSTM.

\subsection{Classification of the composite activities\label{sec:fe_cla}}
\begin{figure}[!h]
    \centering
    \begin{tikzpicture}[node distance = 2.3cm, auto]

        \node [block2, fill=gray!30] (lstm5) {LSTM};

        \node [block2, right of=lstm5] (fc1) {FC1};
        \node [block2, right of=fc1] (fc2) {FC2};
        \node [block2, right of=fc2] (sm1) {SM};


        \path [line] (lstm5) -- (fc1);

        \path [line] (fc1) -- (fc2);
        \path [line] (fc2) -- (sm1);

        
        \node [right=0.17cm of lstm5, yshift=0.3cm] {$\mathbf{s}_{5}$};










        
        
        
        
    \end{tikzpicture}
    \caption{\footnotesize The physical activity is classified using the final internal state ($\mathbf{s}_5$) from the LSTM (see Fig. \ref{fig:comp_feat_ext}). The first fully connected layer (FC1) transforms $s_5$ into a fixed-size representation to suit the classification task. The second fully connected layer (FC2) transforms the output of FC1 into a vector of dimensionality matching the number of supported physical activities. The softmax layer (SM) transforms the output of FC2 into a probability distribution over the physical activities. The classification result is the physical activity with the highest probability.}
    \label{fig:comp_clas}
\end{figure}
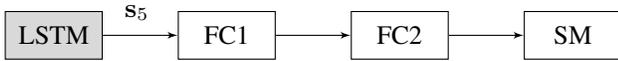

In order to classify the current physical activity, the feature representation of a composite activity ($\mathbf{s}_5$ in Fig. \ref{fig:comp_feat_ext}) is passed through two fully connected layers. The second fully connected layer outputs a vector, with a dimensionality that matches the number of supported physical activities. This vector is then transformed into a probability distribution by a softmax layer, and the classification result is the physical activity that corresponds to the highest relative probability. This process is depicted in Fig. \ref{fig:comp_clas}. The next section describes the setup of the experiments and reports the obtained results.

\subsection{Training the model\label{sec:prop_train}}
Our model is trained using the established backpropagation algorithm (applied recently in \cite{JMLR:v15:srivastava14a,NIPS2014_5423,conf/icml/LiSZ15,NIPS2012_4683}.) The convergence of the backpropagation is affected by the selection of a suitable learning rate ($\alpha$ in Eq. \ref{eq:grad_desc}) \cite{Bengio2012}. We apply Adam algorithm \cite{DBLP:journals/corr/KingmaB14} for adaptive selection of the learning rate (used in \cite{DBLP:conf/nips/ChungKDGCB15,Rasmus:2015:SLL:2969442.2969635,icml2015_gregor15}), which does not require an initial selection of a fixed learning rate. We evaluated empirically that the model converged significantly faster using Adam than with a fixed learning rate (stochastic gradient descent). Adam convergences fast because 1) it calculates a separate learning rate for each of the features in the model and 2) it incorporates momentum in the gradient descent. See \cite{bishop} for details.

The model parametrization used in this paper requires a high number of adjustable parameters (278 696). One challenge with the model is its tendency to overfit the training data; that is, it does not generalize well with previously unseen data \cite{NIPS2012_4824,bishop,JMLR:v15:srivastava14a,Murphy:2012:MLP:2380985}. One indicator of an overfitted model is a large norm of weight parameters \cite{bishop}. If the weight values are large in the inner product (e.g. $\mathbf{W}$ in $\sigma(\mathbf{W}\mathbf{x} + \mathbf{b})$), then small changes in the input cause large changes in the value of the inner product. A model becomes overly specific to the training data with large values of $\mathbf{W}$. To improve the generalization ability, we utilize $L_2$ regularization in the weight values of the fully connected layers, the LSTM and the convolution layers. The $L_2$ regularization penalizes the training of the model with a specified regularization strength ($\gamma$) as $L_2(\mathbf{W},\gamma) = \gamma\sum\nolimits_i W^2_i$ where $\sum\nolimits_i W^2_i$ is the sum of the squared weight values \cite{bishop} and $\mathbf{W} \in \mathbf{\Theta}$. In our experiments, we use the regularization strength of $\gamma=0.01$.


To further improve the generalization ability and to reduce overfitting, we employ a technique called dropout \cite{JMLR:v15:srivastava14a}. We use dropout to randomly disable portions of inputs in the fully connected layers and the LSTM with 50\% probability as in \cite{NIPS2012_4824}. Therefore our model has to learn a robust classification as the model cannot rely on having an access to all of the available information. Dropout forces the computational units to rely on themselves, and not to co-adapt with each other, which improves the classification accuracy of the neural network \cite{NIPS2012_4824, JMLR:v15:srivastava14a}.

Despite the $L_2$ regularization and dropout, we observed overfitting because 1) the representational capacity of the model is high and 2) Adam is an efficient method for stochastic optimization. In the experiments in Section \ref{sec:experiments}, our model consistently provided good results in the beginning of the training procedure. Therefore we utilized an approach called early stopping that trains a model for a limited amount of time. In our experiments, the training was stopped after two iterations of the gradient descent (Eq. \ref{eq:grad_desc}) using Adam algorithm. See \cite{bishop,prechelt_es_tricks,barberBRML2011} for a detailed definition of the early stopping. The following section describes the evaluation experiments and reports their results.

%% file: sections/3rd.tex
\section{Experiments\label{sec:experiments}}
The experiments use data recorded on nine persons. The data originate from the Palantir Context Data Library, see \cite{Parkka:2006,Ermes:2008:DDA:2222942.2223513} for more details. It consists of five minutes of acceleration signal for the following physical activities: walking, Nordic walking, running, soccer, rowing, bicycling, exercise bicycling and lying down. The data were recorded using a triaxial accelerometer worn on the wrist. Two types of experiments are conducted for each test subject: a population system and a personalized system. The resulting classification accuracies are then reported. A population system is created using 40 minutes of acceleration signal (from 8 test subjects) and evaluated using 5 minutes of acceleration signal (from one test subject). This simulates a use-case where a user utilizes a previously created model for human activity detection. A personalized system is constructed for each test subject using the first two minutes of the acceleration signal per physical activity. The classification accuracy is evaluated using the remaining three minutes of the acceleration signal. This simulates a use-case where the user of the system creates a personalized model for human activity detection from scratch.

The following established classifiers are selected as the baseline models in the experiments: logistic regression \cite{bishop}, random forest classifier \cite{Breiman:2001:RF:570181.570182}, decision tree \cite{bishop}, adaptive boosting classifier \cite{Freund+Schapire:1996} and linear support vector machine \cite{bishop}. Gaussian support vector machine is not selected because of its infeasible computation time. The baseline models utilize the data in one go (15 seconds of acceleration signal). We train, evaluate and report the baseline models individually. The next subsection reports the results from the experiments. The baseline models do not model the physical activities as combinations of composite activities. Instead, like in the existing work \cite{6780615,Kwapisz:2011:ARU:1964897.1964918,Parkka:2006,Ermes:2008:DDA:2222942.2223513}, the baseline models attempt to instantaneously determine the current physical activity.

\subsection{Results using a population model\label{sec:exp_res_pop}}
A population system was constructed for each test subject using our proposed model and the baseline models. For example, for the first test subject, a population system was constructed using the data from the remaining (eight) test subjects. The experiments measure the effectiveness of providing a pre-existing, non-personalized model for human activity detection. Table \ref{tab:model_accuracies_pop} reports the mean accuracies for the population systems by repeating their construction and evaluation ten times. The proposed model obtained an overall mean accuracy of $77.91$\%, while the best baseline model (random forest) obtained an overall mean accuracy of $73.52$\%. The proposed model also provides the highest mean accuracy for the individual users. Compared to the the best baseline model, significant improvements in the mean accuracies are obtained using the proposed model for users one ($95.26$\% vs $90.65$\%), three ($95.65$\% vs $91.91$\%) and eight ($81.72$\% vs. $71.53$\%). The results show that our model outperforms the baseline models.

The models do not provide good results for the fifth and the sixth user. This unveils a fundamental problem with the population system; it cannot recognize an activity if a user performs it in a personal, unique, manner. For users five and six, soccer consisted of significant amounts of walking and lying down, and hence the model had trouble distinguishing soccer from the latter two, hence the low accuracies. The following subsection reports results from experiments where the users are provided a personalized system.


\begin{table*}[t!]
    \centering
    \normalsize
    \begin{tabular}{l l l l l l l l l l l}
        & User 1 & User 2 & User 3 & User 4 & User 5 & User 6 & User 7 & User 8 & User 9 & Mean \\
        \hline
        Proposed & \textbf{0.9526} & \textbf{0.9454} & \textbf{0.9565} & \textbf{0.8325} & \textbf{0.3985} & \textbf{0.3862} & \textbf{0.8403} & \textbf{0.8172} & \textbf{0.8824} & \textbf{0.7791} \\
        Logistic regression & 0.5540 & 0.6319 & 0.5549 & 0.6274 & 0.2113 & 0.2916 & 0.5379 & 0.3617 & 0.6215 & 0.4880 \\
        Random forest & 0.9065 & 0.9390 & 0.9191 & 0.8049 & 0.3595 & 0.2973 & 0.8096 & 0.7153 & 0.8659 & 0.7352\\
        Decision tree & 0.8402 & 0.7854 & 0.7946 & 0.6982 & 0.3304 & 0.3089 & 0.6836 & 0.6055 & 0.7235 & 0.6411 \\
        Adaptive boosting & 0.4918 & 0.4602 & 0.5164 & 0.5671 & 0.0925 & 0.3376 & 0.5319 & 0.4115 & 0.4514 & 0.4289 \\
        Linear support vector machine & 0.5361 & 0.5626 & 0.5403 & 0.4033 & 0.1364 & 0.3516 & 0.4405 & 0.3958 & 0.4618 & 0.4254 \\
    \end{tabular}
    \caption{\footnotesize Results using a population system. The cells denote the mean classification accuracies (ten repetitions) for our proposed model and the baseline models. The last column denotes the mean accuracy over all users. The best results are bolded. Our proposed model outperforms the baseline models. Significant benefits are obtained for users 1, 3 and 8 using our proposed model.}
    \label{tab:model_accuracies_pop}
\end{table*}

\subsection{Results using a personalized model\label{sec:exp_res}}
A personalized system was constructed for each test subject using our proposed model and the baseline models. Table \ref{tab:model_accuracies} reports the mean accuracies for the personalized systems by repeating their construction and evaluation ten times. The mean accuracy of our model exceeds $90$\% for the users while the baseline models are accurate for a subset of the users. The overall mean accuracy of the proposed model is $95.28$\%, while the best baseline model (random forest) obtained an overall mean accuracy of $90.03$\%. The proposed model also provides the highest mean accuracy for the individual users. The results show that our model outperforms the baseline models.

As explained above, human activity detection for the user 5 is particularly challenging as physical activities are not well defined. However, our model learns the composite activity of soccer for the user 5, which is one of the advantages of the personalized system. The mean accuracy is $92.03$\% using the proposed model. The baseline models do not model physical activities explicitly as composite activities, which results in significantly lower accuracies than those of our model. The best baseline model obtained a mean accuracy of $76.07$\%. Therefore it is beneficial to use our model when the physical activities are difficult to classify without considering the temporal series of atomic activities.


\begin{table*}[t!]
    \centering
    \normalsize
    \begin{tabular}{l l l l l l l l l l l}
        & User 1 & User 2 & User 3 & User 4 & User 5 & User 6 & User 7 & User 8 & User 9 & Mean \\
        \hline
        Proposed & \textbf{0.9974} & \textbf{0.9784} & \textbf{0.9857} & \textbf{0.9334} & \textbf{0.9203} & \textbf{0.9278} & \textbf{0.9422} & \textbf{0.9065} & \textbf{0.9836} & \textbf{0.9528} \\
        Logistic regression & 0.8602 & 0.8705 & 0.9304 & 0.8435 & 0.6907 & 0.9045 & 0.7999 & 0.6099 & 0.8669 & 0.8196 \\
        Random forest & 0.9964 & 0.8880 & 0.9855 & 0.9059 & 0.7607 & 0.8943 & 0.8523 & 0.8979 & 0.9221 & 0.9003 \\
        Decision tree & 0.9076 & 0.8130 & 0.8939 & 0.8087 & 0.6890 & 0.8412 & 0.7367 & 0.7437 & 0.8470 & 0.8090 \\
        Adaptive boosting & 0.4499 & 0.3207 & 0.5384 & 0.2753 & 0.2215 & 0.2714 & 0.4878 & 0.2707 & 0.7051 & 0.3934 \\
        Linear support vector machine & 0.8690 & 0.8037 & 0.8804 & 0.8145 & 0.5923 & 0.8949 & 0.7896 & 0.6378 & 0.8449 & 0.7919 \\
    \end{tabular}
    \caption{\footnotesize Results using a personalized system. The cells denote the mean classification accuracies (ten repetitions) for our proposed model and the baseline models. The last column denotes the mean accuracy over all users. The best results are bolded. Our proposed model outperforms the baseline models by providing a mean accuracy over 90\% for the users. The baseline models are efficient only for a subset of the users. Compared to the baseline models, users 5, 7 and 9 obtain significantly better results using our proposed model.}
    \label{tab:model_accuracies}
\end{table*}

\section{Importance of the composite activities}
The central assumption of our work is that composite activities are efficiently learned as temporal series of atomic activities. To explicitly study the validity of this assumption, we conducted two experiments in the following two subsections.

\subsection{Human activity detection without composite activities}
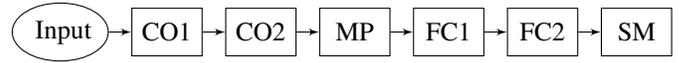
\begin{figure}[!h]
    \centering
    \begin{tikzpicture}[node distance = 1.0cm, auto]
    
        \node [block2_ell, text width=1.9em] (g_input) {Input};
        \node [block2, text width=0.7cm, right=0.3cm of g_input] (fe2) {CO1};
        \node [block2, text width=0.7cm, right=0.3cm of fe2] (fe3) {CO2};
        \node [block2, text width=0.7cm, right=0.3cm of fe3] (fe4) {MP};
        
        \node [block2, text width=0.7cm, right=0.3cm of fe4] (fc1) {FC1};
        \node [block2, text width=0.7cm, right=0.3cm of fc1] (fc2) {FC2};
        \node [block2, text width=0.7cm, right=0.3cm of fc2] (sm1) {SM};


        \path [line] (g_input) -- (fe2);
        \path [line] (fe2) -- (fe3);
        \path [line] (fe3) -- (fe4);
        \path [line] (fe4) -- (fc1);
        \path [line] (fc1) -- (fc2);
        \path [line] (fc2) -- (sm1);
        
    \end{tikzpicture}
    \caption{\footnotesize A trimmed version of the proposed model where components for combining atomic activities (Fig. \ref{fig:comp_feat_ext}) are removed. The resulting model tests the impact on accuracy of not combining atomic activities into composite activities.}
    \label{fig:baseline_no_composite}
\end{figure}

We removed the recurrent components (Fig. \ref{fig:comp_feat_ext}) from the proposed model to test the importance of the composite activities. The trimmed model, which does not explicitly combine the atomic activities in time, is illustrated in Fig. \ref{fig:baseline_no_composite}. The input for the model is 15s of acceleration signal in one go.

The trimmed model was utilized as a population system and a personalized system for the nine users. Table \ref{tab:trimmed_model_accuracy} reports the mean accuracies (ten repetitions) for the trimmed model, as well as the full model, for comparison. The full model provides a better performance for the users with mean accuracies of 77.9\% in the population system and 95.3\% in the personalized system. The respective figures for the trimmed model are $71.52$\% and $91.22$\%, respectively. The results suggest that the utilization of composite activities improves the accuracy of human activity detection.

\begin{table*}[t!]
    \centering
    \normalsize
    \begin{tabular}{l l l l l l l l l l l}
        & User 1 & User 2 & User 3 & User 4 & User 5 & User 6 & User 7 & User 8 & User 9 & Mean \\
        \hline
        Population (full) & \textbf{0.9526} & \textbf{0.9454} & \textbf{0.9565} & \textbf{0.8325} & \textbf{0.3985} & \textbf{0.3862} & \textbf{0.8403} & \textbf{0.8172} & \textbf{0.8824} & \textbf{0.7791} \\
        Population (trimmed) & 0.8290 & 0.8997 & 0.8546 & 0.8204 & 0.2896 & 0.3829 & 0.8257 & 0.7234 & 0.8112 & 0.7152 \\
        \hline
        Personalized (full) & \textbf{0.9974} & \textbf{0.9784} & \textbf{0.9857} & \textbf{0.9334} & \textbf{0.9203} & \textbf{0.9278} & \textbf{0.9422} & \textbf{0.9065} & \textbf{0.9836} & \textbf{0.9528} \\
        Personalized (trimmed) & 0.9662 & 0.9222 & 0.9775 & 0.8985 & 0.7898 & 0.9120 & 0.9405 & 0.8255 & 0.9776 & 0.9122 \\
    \end{tabular}
    \caption{\footnotesize Population and personalized system results for the trimmed and full version of the proposed model. The trimmed version does not combine atomic activities in time. The cells denote the mean classification accuracies (ten repetitions). The last column denotes the mean accuracy over all users. The best results are bolded. Compared to the trimmed version of the model, the full version provides a higher accuracy for the users. The results suggest that it is beneficial for human activity detection to combine atomic activities into composite activities.}
    \label{tab:trimmed_model_accuracy}
\end{table*}

\subsection{Visualization of the recurrent states}
\begin{figure*}[t!]
    \centering
    \begin{subfigure}[b]{3.3in}
        \centering
        \includegraphics[width=3.3in]{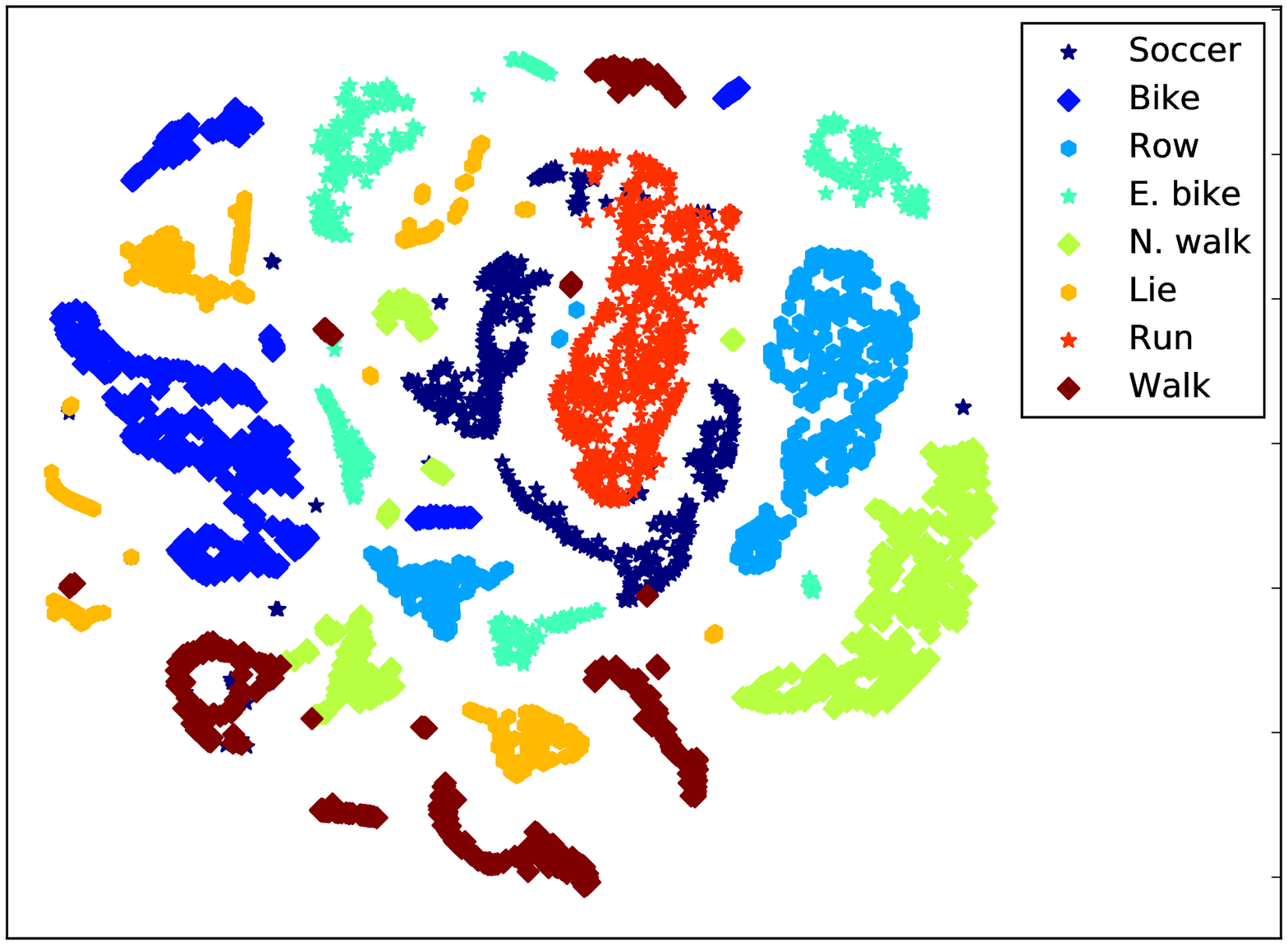}
    \end{subfigure}
    \begin{subfigure}[b]{3.3in}
        \centering
        \includegraphics[width=3.3in]{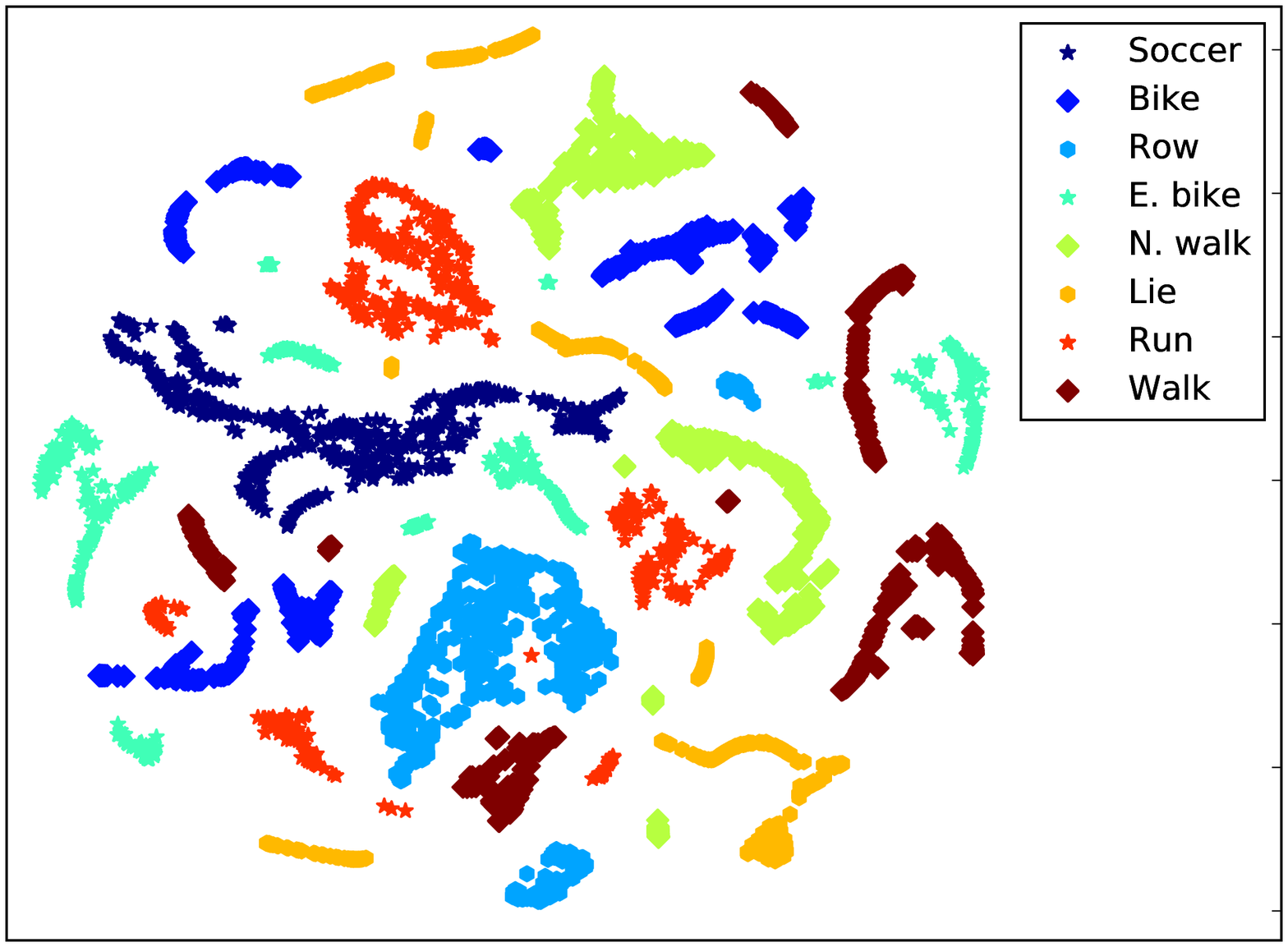}
    \end{subfigure}
    \caption{\footnotesize Visualization of the internal states of the LSTM ($\mathbf{s}_i$ in Fig. \ref{fig:comp_feat_ext}) per physical activity for the second test subject. Best viewed in color. The internal states are projected in a two-dimensional manifold using t-SNE algorithm. \textbf{Left}: The first internal state ($\mathbf{s}_1$ in Fig. \ref{fig:comp_feat_ext}.) \textbf{Right}: The final internal state ($\mathbf{s}_5$ in in Fig. \ref{fig:comp_feat_ext}.) Notice that the composite activities can overlap with related atomic activities (e.g. soccer, walking and running overlap) in the first internal state. However, using $\mathbf{s}_5$, all of the physical activities are separated from each other and the clusters are more compact.}
    \label{fig:visualize_lstm}
\end{figure*}

We projected the first and the last internal states of the LSTM ($\mathbf{s}_1$ and $\mathbf{s}_5$ in Fig. \ref{fig:comp_feat_ext}) into a two-dimensional manifold. The projection was computed using t-distributed stochastic neighbor embedding (t-SNE, \cite{maaten:tsne:12534106,VanDerMaaten:2014:ATU:2627435.2697068}), which is designed to visualize high-dimensional data. The internal LSTM states of a personalized model for the second test subject were computed using three minutes of validation data. Fig. \ref{fig:visualize_lstm} illustrates the obtained results using t-SNE where the left image is the first internal state and the right image is the last internal state. Notice that initially some of the composite activities are mixed with atomic activities. For example, soccer overlaps with walking and running. This suggests that it is a fundamentally challenging task to instantaneously recognize composite activities. However, walking and running do not overlap with soccer in the last internal state of the LSTM. Therefore our model has learned to identify composite activities as temporal series of atomic activities.


%% file: sections/4th.tex
\section{Discussion and conclusion}
There are complex activities (composite activities) that consist of simpler activities (atomic activities). It is a fundamentally challenging task to classify a composite activity without decomposing it into atomic activities. In this paper, we have proposed a model to automatically learn the atomic activities and how to combine them into composite activities. The experiments in Section \ref{sec:experiments} show the following benefits of our model:

\begin{enumerate}
    \item Our proposed model learns to form a set of atomic activities and to combine them meaningfully. Instantaneous classification is not sufficient and one needs to learn the composite activities as combinations of atomic activities over time. This is suggested from inspecting Fig. \ref{fig:visualize_lstm} and the results in Table \ref{tab:trimmed_model_accuracy}.
    \item The proposed model outperforms the baseline models by obtaining an overall mean accuracy of $77.91$\% (population) and $95.28$\% (personalized). The corresponding accuracies of the best baseline model were $73.52$\% and $90.03$\%. However, the corresponding accuracies of the proposed model without utilizing the composite activities was $71.52$\% and $91.22$\%.
\end{enumerate}

The limitations of our model are the utilization of 15s of acceleration signal and the memory consumption. Depending on the application domain, 15 seconds can be adequate. However, our model is not suited for time-critical systems that require a low latency for the results. Additionally, the current parametrization of our model uses 8.92Mb (32bit floating point values) of memory. This can be a large amount of memory for small embedded devices. One of our goals is to find a small parametrization, which retains a good accuracy. However, we believe that the amount of available device memory is going to increase. Future work includes the following:
\begin{itemize}
    \item Further experiments using different parametrizations of the proposed model.
    \item Unsupervised segmentation of acceleration signal using the learned features in Fig. \ref{fig:comp_feat_ext} and \ref{fig:comp_clas}.
    \item Fusion of different sensor signals for human activity detection.
\end{itemize}

%% file: main.bbl
\begin{thebibliography}{10}
\providecommand{\url}[1]{#1}
\csname url@samestyle\endcsname
\providecommand{\newblock}{\relax}
\providecommand{\bibinfo}[2]{#2}
\providecommand{\BIBentrySTDinterwordspacing}{\spaceskip=0pt\relax}
\providecommand{\BIBentryALTinterwordstretchfactor}{4}
\providecommand{\BIBentryALTinterwordspacing}{\spaceskip=\fontdimen2\font plus
\BIBentryALTinterwordstretchfactor\fontdimen3\font minus
  \fontdimen4\font\relax}
\providecommand{\BIBforeignlanguage}[2]{{%
\expandafter\ifx\csname l@#1\endcsname\relax
\typeout{** WARNING: IEEEtran.bst: No hyphenation pattern has been}%
\typeout{** loaded for the language `#1'. Using the pattern for}%
\typeout{** the default language instead.}%
\else
\language=\csname l@#1\endcsname
\fi
#2}}
\providecommand{\BIBdecl}{\relax}
\BIBdecl

\bibitem{973004}
J.~M{\"a}ntyj{\"a}rvi, J.~Himberg, and T.~Sepp{\"a}nen, ``Recognizing human
  motion with multiple acceleration sensors,'' in \emph{Proceedings of the 2001
  IEEE International Conference on Systems, Man, and Cybernetics}, vol.~2,
  2001, pp. 747--752.

\bibitem{mathie-classify-daily-movements}
M.~Mathie, B.~Celler, N.~Lovell, and A.~Coster, ``Classification of basic daily
  movements using a triaxial accelerometer,'' \emph{Medical and Biological
  Engineering and Computing}, vol.~42, no.~5, pp. 679--687, 2004.

\bibitem{Karantonis:2006:IRH:2222932.2223358}
D.~M. Karantonis, M.~R. Narayanan, M.~Mathie, N.~H. Lovell, and B.~G. Celler,
  ``Implementation of a real-time human movement classifier using a triaxial
  accelerometer for ambulatory monitoring,'' \emph{IEEE Transactions on
  Information Technology in Biomedicine}, vol.~10, no.~1, pp. 156--167, Jan.
  2006.

\bibitem{6780615}
P.~Gupta and T.~Dallas, ``Feature selection and activity recognition system
  using a single triaxial accelerometer,'' \emph{IEEE Transactions on
  Biomedical Engineering}, vol.~61, no.~6, pp. 1780--1786, June 2014.

\bibitem{Khan:2010:TAP:1870372.1870376}
A.~Khan, Y.~Lee, S.~Lee, and T.~Kim, ``A triaxial accelerometer-based
  physical-activity recognition via augmented-signal features and a
  hierarchical recognizer,'' \emph{IEEE Transactions on Information Technology
  in Biomedicine}, vol.~14, no.~5, pp. 1166--1172, Sep 2010.

\bibitem{chernbumroong-activity}
S.~Chernbumroong, A.~Atkins, and H.~Yu, ``Activity classification using a
  single wrist-worn accelerometer,'' in \emph{Proceedings of the 2011 5th
  International Conference on Software, Knowledge Information, Industrial
  Management and Applications (SKIMA)}, Sept 2011, pp. 1--6.

\bibitem{godfrey-activity-class}
A.~Godfrey, A.~Bourke, G.~Olaighin, P.~van~de Ven, and J.~Nelson, ``Activity
  classification using a single chest mounted tri-axial accelerometer,''
  \emph{Medical Engineering \& Physics}, vol.~33, pp. 1127--1135, 2011.

\bibitem{Kwapisz:2011:ARU:1964897.1964918}
J.~Kwapisz, G.~Weiss, and S.~Moore, ``Activity recognition using cell phone
  accelerometers,'' \emph{SIGKDD Explor. Newsl.}, vol.~12, no.~2, pp. 74--82,
  Mar 2011.

\bibitem{6217313}
J.~Petersen, D.~Austin, R.~Sack, and T.~Hayes, ``Actigraphy-based scratch
  detection using logistic regression,'' \emph{IEEE Journal of Biomedical and
  Health Informatics}, vol.~17, no.~2, pp. 277--283, Mar 2013.

\bibitem{6399499}
W.~Cheng and D.~Jhan, ``Triaxial accelerometer-based fall detection method
  using a self-constructing cascade-adaboost-svm classifier,'' \emph{IEEE
  Journal of Biomedical and Health Informatics}, vol.~17, no.~2, pp. 411--419,
  Mar 2013.

\bibitem{6346377}
A.~Matic, V.~Osmani, and O.~Mayora, ``Speech activity detection using
  accelerometer,'' in \emph{Proceedings of the 2012 IEEE Annual International
  Conference on Engineering in Medicine and Biology Society}, Aug 2012, pp.
  2112--2115.

\bibitem{Parkka:2006}
J.~P{\"a}rkk{\"a}, M.~Ermes, P.~Korpip{\"a}{\"a}, J.~M{\"a}ntyj{\"a}rvi,
  J.~Peltola, and I.~Korhonen, ``Activity classification using realistic data
  from wearable sensors,'' \emph{IEEE Transactions on Information Technology in
  Biomedicine}, vol.~10, no.~1, pp. 119--128, Jan. 2006.

\bibitem{Ermes:2008:DDA:2222942.2223513}
M.~Ermes, J.~P{\"a}rkk{\"a}, J.~M{\"a}ntyj{\"a}rvi, and I.~Korhonen,
  ``Detection of daily activities and sports with wearable sensors in
  controlled and uncontrolled conditions,'' \emph{IEEE Transactions on
  Information Technology in Biomedicine}, vol.~12, no.~1, pp. 20--26, Jan.
  2008.

\bibitem{Schmidhuber201585}
J.~Schmidhuber, ``Deep learning in neural networks: An overview,'' \emph{Neural
  Networks}, vol.~61, pp. 85--117, 2015.

\bibitem{Koller:2009:PGM:1795555}
D.~Koller and N.~Friedman, \emph{Probabilistic Graphical Models: Principles and
  Techniques - Adaptive Computation and Machine Learning}.\hskip 1em plus 0.5em
  minus 0.4em\relax The MIT Press, 2009.

\bibitem{Bastien-Theano-2012}
F.~Bastien, P.~Lamblin, R.~Pascanu, J.~Bergstra, I.~Goodfellow, A.~Bergeron,
  N.~Bouchard, and Y.~Bengio, ``Theano: new features and speed improvements,''
  in \emph{Deep Learning and Unsupervised Feature Learning NIPS 2012 Workshop},
  2012.

\bibitem{bishop}
C.~Bishop, \emph{Pattern Recognition and Machine Learning}.\hskip 1em plus
  0.5em minus 0.4em\relax Secaucus, NJ, USA: Springer-Verlag New York, Inc.,
  2006.

\bibitem{Breiman:2001:RF:570181.570182}
L.~Breiman, ``Random forests,'' \emph{Machine learning}, vol.~45, no.~1, pp.
  5--32, Oct 2001.

\bibitem{Freund+Schapire:1996}
Y.~Freund and R.~Schapire, ``Experiments with a new boosting algorithm,'' in
  \emph{Proceedings of the Thirteenth International Conference on Machine
  Learning (ICML 1996)}, L.~Saitta, Ed.\hskip 1em plus 0.5em minus 0.4em\relax
  Morgan Kaufmann, 1996, pp. 148--156.

\bibitem{Hochreiter:1997:LSM:1246443.1246450}
S.~Hochreiter and J.~Schmidhuber, ``Long short-term memory,'' \emph{Neural
  Computing}, vol.~9, no.~8, pp. 1735--1780, Nov 1997.

\bibitem{deep_learning_nature_hinton}
Y.~LeCun, Y.~Bengio, and G.~Hinton, ``Deep learning,'' \emph{Nature}, pp.
  436--444, 2015.

\bibitem{NIPS2012_4824}
A.~Krizhevsky, I.~Sutskever, and G.~Hinton, ``Imagenet classification with deep
  convolutional neural networks,'' in \emph{Advances in Neural Information
  Processing Systems 25}, F.~Pereira, C.~Burges, L.~Bottou, and K.~Weinberger,
  Eds.\hskip 1em plus 0.5em minus 0.4em\relax Curran Associates, Inc., 2012,
  pp. 1097--1105.

\bibitem{Bengio2012}
Y.~Bengio, \emph{Neural Networks: Tricks of the Trade: Second Edition}.\hskip
  1em plus 0.5em minus 0.4em\relax Berlin, Heidelberg: Springer Berlin
  Heidelberg, 2012, ch. Practical Recommendations for Gradient-Based Training
  of Deep Architectures, pp. 437--478.

\bibitem{AISTATS2011_GlorotBB11}
X.~Glorot, A.~Bordes, and Y.~Bengio, ``Deep sparse rectifier neural networks,''
  in \emph{Proceedings of the Fourteenth International Conference on Artificial
  Intelligence and Statistics}, vol.~15, 2011, pp. 315--323.

\bibitem{Theodoridis:2008:PRF:1457541}
S.~Theodoridis and K.~Koutroumbas, \emph{Pattern Recognition}, 4th~ed.\hskip
  1em plus 0.5em minus 0.4em\relax Academic Press, 2008.

\bibitem{Mitchell:1997:ML:541177}
T.~Mitchell, \emph{Machine Learning}, 1st~ed.\hskip 1em plus 0.5em minus
  0.4em\relax New York, NY, USA: McGraw-Hill, Inc., 1997.

\bibitem{JMLR:v15:srivastava14a}
N.~Srivastava, G.~Hinton, A.~Krizhevsky, I.~Sutskever, and R.~Salakhutdinov,
  ``Dropout: A simple way to prevent neural networks from overfitting,''
  \emph{Journal of Machine Learning Research}, vol.~15, pp. 1929--1958, 2014.

\bibitem{NIPS2014_5423}
I.~Goodfellow, J.~Pouget-Abadie, M.~Mirza, B.~Xu, D.~Warde-Farley, S.~Ozair,
  A.~Courville, and Y.~Bengio, ``Generative adversarial nets,'' in
  \emph{Advances in Neural Information Processing Systems 27}.\hskip 1em plus
  0.5em minus 0.4em\relax Curran Associates, Inc., 2014, pp. 2672--2680.

\bibitem{conf/icml/LiSZ15}
Y.~Li, K.~Swersky, and R.~Zemel, ``Generative moment matching networks,'' in
  \emph{ICML}, ser. JMLR Proceedings, F.~Bach and D.~Blei, Eds., vol.~37, 2015,
  pp. 1718--1727.

\bibitem{NIPS2012_4683}
N.~Srivastava and R.~Salakhutdinov, ``Multimodal learning with deep boltzmann
  machines,'' in \emph{Advances in Neural Information Processing Systems
  25}.\hskip 1em plus 0.5em minus 0.4em\relax Curran Associates, Inc., 2012,
  pp. 2222--2230.

\bibitem{DBLP:journals/corr/KingmaB14}
D.~Kingma and J.~Ba, ``Adam: {A} method for stochastic optimization,''
  \emph{CoRR}, vol. abs/1412.6980, 2014.

\bibitem{DBLP:conf/nips/ChungKDGCB15}
J.~Chung, K.~Kastner, L.~Dinh, K.~Goel, A.~Courville, and Y.~Bengio, ``A
  recurrent latent variable model for sequential data,'' in \emph{Advances in
  Neural Information Processing Systems 28: Annual Conference on Neural
  Information Processing}, 2015, pp. 2980--2988.

\bibitem{Rasmus:2015:SLL:2969442.2969635}
A.~Rasmus, H.~Valpola, M.~Honkala, M.~Berglund, and T.~Raiko, ``Semi-supervised
  learning with ladder networks,'' in \emph{Proceedings of the 28th
  International Conference on Neural Information Processing Systems}, ser.
  NIPS'15, 2015, pp. 3546--3554.

\bibitem{icml2015_gregor15}
K.~Gregor, I.~Danihelka, A.~Graves, D.~Rezende, and D.~Wierstra, ``Draw: A
  recurrent neural network for image generation,'' in \emph{Proceedings of the
  32nd International Conference on Machine Learning (ICML-15)}.\hskip 1em plus
  0.5em minus 0.4em\relax JMLR Workshop and Conference Proceedings, 2015, pp.
  1462--1471.

\bibitem{Murphy:2012:MLP:2380985}
K.~P. Murphy, \emph{Machine Learning: A Probabilistic Perspective}.\hskip 1em
  plus 0.5em minus 0.4em\relax The MIT Press, 2012.

\bibitem{prechelt_es_tricks}
L.~Prechelt, ``Early stopping — but when?'' in \emph{Neural Networks: Tricks
  of the Trade}, ser. Lecture Notes in Computer Science, 2012, vol. 7700.

\bibitem{barberBRML2011}
D.~Barber, \emph{Bayesian Reasoning and Machine Learning}.\hskip 1em plus 0.5em
  minus 0.4em\relax New York, NY, USA: Cambridge University Press, 2012.

\bibitem{maaten:tsne:12534106}
L.~Van Der~Maaten and G.~Hinton, ``Visualizing high-dimensional data using
  t-{SNE},'' \emph{Journal of Machine Learning Research}, vol.~9, pp.
  2579--2605, Nov 2008.

\bibitem{VanDerMaaten:2014:ATU:2627435.2697068}
L.~Van Der~Maaten, ``Accelerating t-{SNE} using tree-based algorithms,''
  \emph{Journal of Machine Learning Research}, vol.~15, no.~1, pp. 3221--3245,
  Jan 2014.

\end{thebibliography}
